\documentclass[runningheads]{llncs}
\usepackage[review]{}

\usepackage{eccvabbrv}

\usepackage{graphicx}
\usepackage{booktabs}
\usepackage{multirow}
\usepackage{amssymb}
\newcounter{subtable}[table]

\newcommand{\subcaption}[1]{\refstepcounter{subtable}\par\smallskip\noindent\textbf{(\alph{subtable}) #1}\par\smallskip}

\usepackage{hyperref}

\usepackage{orcidlink}

\begin{document}

\title{CHASMBrain: \textbf{C}oarse-to-fine \textbf{H}ierarchical \textbf{A}rchitecture with \textbf{S}equential \textbf{M}amba for \textbf{Brain} Reconstruction}

\titlerunning{CHASM Brain}

\author{Hoang-Son Vo
  \inst{1} $^\dagger$ \orcidlink{0009-0001-3278-727X}
  \and Van-Hung Bui \inst{1} $^\dagger$ \orcidlink {0009-0005-1376-6616}
  \and Minh-Huy Mai-Duc \inst{2}
  \and Tien-Dung Mai \inst{3}\orcidlink{2222--3333-4444-5555}
  \and Soo-Hyung Kim \inst{1} $^*$ \orcidlink{0000-0003-3575-5035}
}

\authorrunning{Son et al.}

\institute{Chonnam National University, Gwangju, Republic of Korea
  \and Vietnam National University - Ho Chi Minh City, University of Science, Vietnam
  \and Institute for Cybersecurity and Digital Technologies, Russia
  $^\dagger$ These authors contributed equally to this work.\\
  $^*$ Corresponding author
}

\maketitle

\begin{abstract}
Understanding the relationship between deep visual representations and the human visual system is a fundamental challenge in computational neuroscience. While modern vision models achieve strong performance in image recognition, their correspondence with the hierarchical organization of the human visual cortex remains an open question. In this study, we propose CHASMBrain, a novel hierarchical two-stage framework for image-to-fMRI encoding. Our architecture leverages a dual-stream Mamba design to explicitly separate and process global semantic tokens and local spatial patches, motivated by the functional organization of the visual cortex. A coarse-to-fine strategy is employed: Stage 1 predicts denoised ROI-level activations, while Stage 2 refines these coarse responses into full voxel-level predictions using a Mamba-VAE. Experiments on the Natural Scenes Dataset (NSD) demonstrate that our method achieves a Pearson correlation of 0.429 and an MSE of 0.261, outperforming all evaluated baselines including ridge regression and DINOv2 linear probes. Beyond predictive performance, causal branch-ablation experiments reveal an asymmetric specialization: the patch stream is specifically locked to early visual cortex (retinotopic regions), while the CLS stream contributes broader semantic context to higher-order areas---a correspondence that holds causally, not merely correlationally. Cross-subject transfer experiments further show that the learned backbone generalizes across individuals with minimal per-subject adaptation, suggesting the model captures a shared, subject-agnostic visual representation.


\end{abstract}

\section{Introduction}
\label{sec:intro}

Vision is the primary sensory channel through which humans acquire information about the external world. Even a single second of natural visual input carries a staggering amount of information. To understand how the brain parses this input, computational neuroscience relies on visual encoding models that describe neural responses as functions of stimulus features. Functional magnetic resonance imaging (fMRI) is central to this effort: by measuring blood-oxygen-level-dependent signals, fMRI provides a non-invasive way to map visual stimuli to localized responses in the human visual cortex.

Historically, encoding models depended on hand-crafted features. Early work used mathematical constructs such as Gabor filters to approximate sensitivity to edges, orientations, and contrast. As naturalistic vision benchmarks became more challenging, the field shifted from manual descriptors to learning-based approaches powered by deep neural networks, which extract rich data-driven representations directly from images.

This transition also revealed a strong correspondence between artificial hierarchies and biological vision. Prior studies show that layers of deep neural networks (DNNs) map onto successive processing stages in human visual cortex \cite{ootaDeepNeuralNetworks2024}. Early convolutional layers are predictive of activity in low-level areas such as V1 and V2 \cite{wangNeuralTaskonomyInferring2019}, while deeper layers better align with higher-level regions, including inferior temporal (IT) cortex and fusiform face area (FFA), that support complex object recognition \cite{luMultimodalFoundationModels2022}. These findings suggest shared computational principles across artificial and biological systems.

To better capture this organization, recent models explicitly separate local and global representations. Early visual cortex is retinotopic, making it especially sensitive to local, patch-level structure. Higher-level regions encode more holistic semantics that reflect object identity and behavioral relevance, beyond low-level geometry \cite{castelloEncodingModelsFunctional}. Disentangling local spatial pooling (``where'') from global feature tuning (``what'') has therefore become a key design principle, as shown by frameworks such as banded ridge regression and targeted transformer queries \cite{stFeatureweightedReceptiveField2018,adeliPredictingBrainActivity2023}.

This separability is consistent with evidence for structural similarity between deep visual representations and the human visual system. From striate to extrastriate cortex, representational complexity increases along a spatial hierarchy \cite{eickenbergSeeingItAll2017}. Deep models exhibit a comparable layer-wise progression \cite{schrimpfBrainScoreWhichArtificial2018}, often described as a form of hierarchical isomorphism between model stages and anatomical pathways spanning ventral and dorsal streams \cite{wenNeuralEncodingDecoding2018,linStackedRegressionsStructured2024}.

Building on these insights, this paper presents an interpretable, biologically inspired dual-stream framework for neural visual encoding. While prior deep models often achieve high predictive accuracy, they can remain difficult to interpret. The proposed framework incorporates cortical priors by separating spatial location parameters from feature-weight parameters, enabling direct inspection of receptive-field structure \cite{stFeatureweightedReceptiveField2018,linStackedRegressionsStructured2024}. It further mirrors neuroanatomical specialization by decoupling spatial detail from semantic context, following evidence from brain-like hierarchical processing \cite{kubiliusBrainLikeObjectRecognition2019}. This design aims to improve both predictive performance and scientific interpretability.

To bridge the gap between artificial visual representations and human neural activity, the core contributions of this work are summarized as follows:

\begin{itemize}

  \item \textbf{CHASMBrain}, a hierarchical two-stage framework for image-to-fMRI encoding that achieves a Pearson correlation of 0.429 and an MSE of 0.261 on NSD, outperforming not only recent deep encoding models (SynBrain, MindSimulator) but also strong simple baselines including ridge regression and DINOv2 layer-wise linear probes.

  \item \textbf{A sequential coarse-to-fine learning mechanism} specifically designed to overcome the challenge of high intrinsic noise in fMRI signals, by separating robust semantic alignment at the ROI level from fine-grained voxel-level refinement via a Mamba-VAE.

  \item \textbf{Causal stream specialization}: branch-ablation and input-routing experiments show that the patch (local) stream is specifically locked to early visual cortex---removing it hurts early regions far more than higher regions---while the CLS (global) stream contributes broader semantic context. Routing streams to mismatched cortical targets (``Swap'') causes a $1.9\times$ asymmetric drop in early-region Pearson, confirming the correspondence is causal rather than post-hoc.

  \item \textbf{Cross-subject generalizability}: a backbone trained on source subjects and frozen, with only a lightweight per-subject head fine-tuned ($\leq$30 epochs), reaches near within-subject performance on held-out subjects, suggesting the learned representation is subject-agnostic at the backbone level.

\end{itemize}

\section{Proposed Method}

\begin{figure}[ht]
  \centering
  \includegraphics[width=\textwidth]{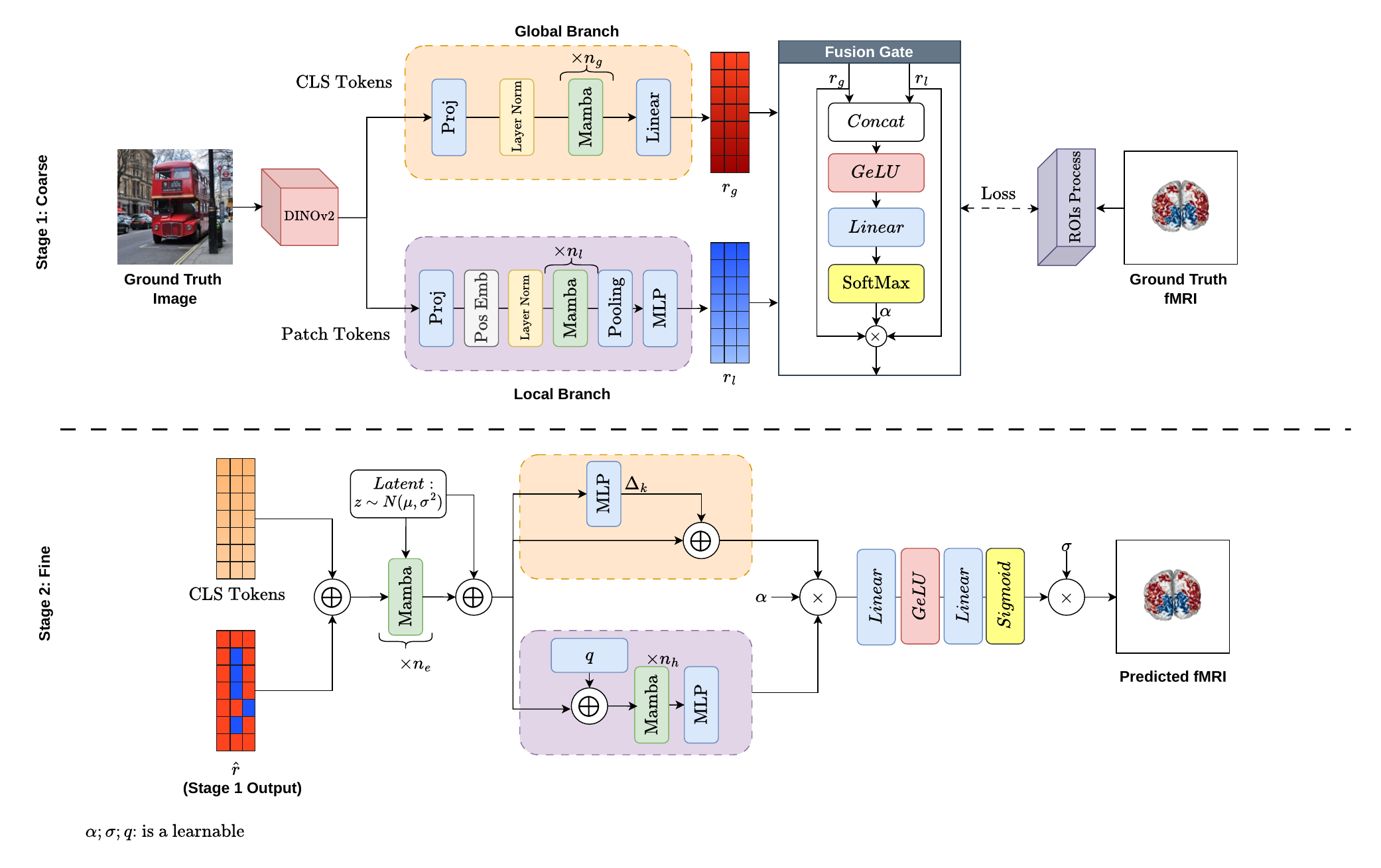}
  \caption{Overview of the proposed hierarchical two-stage framework for image-to-fMRI reconstruction.
    \textbf{Stage 1 (Coarse)}: A dual-stream Mamba architecture processes global (CLS token) and local (patch tokens) visual features separately, then fuses them via an adaptive gate to predict ROI-level activations.
  \textbf{Stage 2 (Fine)}: A Mamba-VAE architecture takes the visual features and ROI predictions as input, encodes them through Mamba blocks, samples from a learned latent space, and decodes to full voxel-level fMRI predictions through parallel global and local branches.}
  \label{fig:architecture}
\end{figure}

We propose a hierarchical two-stage framework for image-to-fMRI reconstruction, as shown in Figure~\ref{fig:architecture}. Stage 1 first predicts denoised responses at the ROI (Region of Interest) level, and Stage 2 then refines these coarse responses into full voxel-level predictions. This design separates robust semantic alignment from fine-grained signal recovery, which makes the overall learning process more stable. Crucially, Stage 1's dual-stream architecture extracts and encodes local spatial details into ROI representations, allowing Stage 2 to focus on refinement using these encoded structures rather than re-extracting features from raw patches.

\subsection{Stage 1: Dual-Stream Mamba for Coarse Prediction}

Stage 1 learns ROI-level neural responses from visual inputs by combining complementary cues: global semantics and local spatial details. To this end, we use a dual-stream architecture where each stream specializes in one type of visual representation and both streams are fused adaptively.

\subsubsection{Visual Feature Extraction}

Given an input image $I$, we extract visual features using a pretrained DINOv2 encoder~\cite{oquab2023dinov2}. The encoder outputs two representations:
\begin{itemize}
  \item \textbf{CLS token}: a single vector summarizing global image semantics.
  \item \textbf{Patch tokens}: a sequence of vectors preserving local spatial structure.
\end{itemize}
We choose DINOv2 because it is pretrained via self-supervised learning, which is effective for capturing fine visual details and structural patterns. This property is well aligned with our hypothesis that both detailed local structure and global visual semantics are critical for accurate image-to-fMRI reconstruction.

\subsubsection{ROI Aggregation for Noise Reduction}

Directly regressing all voxels (e.g., 15,724 dimensions) is difficult because fMRI signals are high-dimensional and noisy, with variance from physiological artifacts, scanner noise, and trial-level fluctuations. Therefore, Stage 1 predicts ROI-level targets as an intermediate denoising step.

We partition voxels into $K$ non-overlapping clusters, each containing $V$ spatially adjacent voxels. Compared with anatomical parcellations such as Schaefer~\cite{schaefer2018local}, our sequential clustering provides flexible granularity tailored to the target voxel space.

For each cluster, instead of averaging all voxels, we aggregate only the top-$p$\% activations:
\begin{equation}
  r_k = \frac{1}{|S_k|} \sum_{i \in S_k} v_i
  \quad \mbox{where} \quad S_k = \mathrm{TopK}(\{v_i\}_{i \in C_k}, \lfloor p \cdot V \rfloor)
\end{equation}
where $r_k$ is the aggregated value of ROI $k$, $C_k$ is the voxel index set of cluster $k$, and $\mathrm{TopK}$ returns the highest-activation subset. This strategy preserves informative peaks while reducing low-activation noise; it is especially useful after z-score normalization, where naive mean aggregation can collapse toward zero.

\subsubsection{Global Branch}

The Global branch models high-level semantics from the CLS token. The projected CLS features are treated as a sequence across the Mamba block ~\cite{mamba2} depth, allowing selective state-space modeling to filter task-irrelevant dimensions progressively

We adopt Mamba primarily because its selective state-space mechanism can focus on task-relevant information while suppressing irrelevant signals, which aligns well with brain activity prediction.

\subsubsection{Local Branch}

The Local branch models fine-grained structure from patch tokens. We first project patch tokens, add learnable positional embeddings, and apply normalization. The sequence is then processed by a deeper stack of Mamba blocks than in the Global branch, reflecting the larger token sequence (256 patches). Finally, mean pooling and a two-layer GELU MLP produce ROI-level predictions.

\subsubsection{Adaptive Fusion Gate}

To combine the two streams, we use a learnable fusion gate rather than fixed coefficients. The gate takes concatenated global/local features, applies a two-layer GELU MLP, and outputs softmax-based fusion weights. The final ROI prediction is
\begin{equation}
  \hat{\mathbf{r}} = \alpha \cdot \mathbf{r}_{g} + (1 - \alpha) \cdot \mathbf{r}_{l},
\end{equation}
where $\mathbf{r}_{g}$ and $\mathbf{r}_{l}$ are branch outputs and $\alpha \in [0,1]$ is learned per input. This mechanism lets the model emphasize global context for simple scenes and local details for complex scenes.

\subsubsection{Contrastive Alignment}

Beyond regression, we encourage representation-level alignment between image features and brain responses via a CLIP-style contrastive objective~\cite{pmlr-v139-radford21a}. Specifically, fused visual features and ground-truth ROI activations are projected into a shared embedding space using separate two-layer GELU projection heads with layer normalization. An InfoNCE loss then pulls matched image--brain pairs together while pushing mismatched pairs apart.

\subsubsection{Training Objective}

The Stage 1 objective combines pointwise regression, correlation alignment, and contrastive alignment:
\begin{equation}
  \mathcal{L}_{stage1} = \mathcal{L}_{MSE} + \gamma \cdot \mathcal{L}_{Pearson} + \lambda \cdot \mathcal{L}_{InfoNCE}.
\end{equation}
Here, $\mathcal{L}_{MSE}$ minimizes amplitude error between predicted and ground-truth ROI activations, $\mathcal{L}_{Pearson}$ encourages high correlation between their response patterns, and $\mathcal{L}_{InfoNCE}$ improves cross-modal discriminability. Hyperparameters $\gamma$ and $\lambda$ balance these three terms.

\subsection{Stage 2: Mamba-VAE for Fine Prediction}

Stage 2 maps coarse ROI predictions to full voxel resolution. We design this stage as a Mamba-VAE to jointly model structured dependencies and trial-to-trial variability in fMRI responses.

\subsubsection{Input Fusion}

Stage 2 receives two inputs: visual features from DINOv2 and ROI predictions from Stage 1 ($\hat{\mathbf{r}}$). We first project visual features (CLS token concatenated with pooled patch features) into a common embedding space. Notably, we use mean-pooled patch features rather than individual patch tokens, as Stage 1's Local branch has already extracted and encoded fine-grained spatial details into the ROI representations. This avoids redundant feature extraction and reduces computational overhead while maintaining spatial specificity through the ROI structure. Each ROI value is then tokenized and augmented with learnable positional embeddings to preserve region order. Concatenating visual and ROI tokens forms a unified sequence encoding both stimulus content and coarse neural priors.

\subsubsection{Mamba Encoder}

A stack of Mamba blocks encodes the fused sequence and learns interactions between visual evidence and predicted neural structure. This encoder produces latent token representations that capture global image--brain dependencies before voxel-level decoding.

\subsubsection{Variational Latent Space}

Because fMRI responses vary even for repeated presentations of the same stimulus, we model uncertainty with a variational latent space. After sequence encoding, pooled features are passed to a VAE head:
\begin{equation}
  \mu, \log\sigma^2 = \mathrm{VAEEncoder}(\mathrm{Pool}(\mathbf{h})).
\end{equation}
During training, we sample
\begin{equation}
  \mathbf{z} = \mu + \sigma \odot \epsilon, \quad \epsilon \sim \mathcal{N}(\mathbf{0}, \mathbf{I}),
\end{equation}
and inject $\mathbf{z}$ back into token representations. During inference, we use $\mu$ for deterministic prediction.

\subsubsection{Feature Split for Dual Decoding}

After latent injection, we obtain a shared token sequence $\tilde{\mathbf{H}} \in \mathbb{R}^{(K+1)\times d}$, including one visual token and $K$ ROI tokens. We then split features by token role (not by separate encoders): the full sequence is routed to a holistic decoder, while ROI tokens are selected and routed to an ROI-wise residual decoder. Formally,
\begin{equation}
  \tilde{\mathbf{H}}_{global} = \tilde{\mathbf{H}}, \quad \tilde{\mathbf{H}}_{local} = \tilde{\mathbf{H}}_{2:K+1}.
\end{equation}
This token-level split allows both decoders to share the same encoded context while specializing in complementary prediction behaviors.

\subsubsection{Holistic Voxel Decoder}

The Holistic Voxel Decoder predicts the full voxel vector jointly from the global context. We append learnable query tokens $\mathbf{q} \in \mathbb{R}^{M \times d}$ to the latent-injected sequence, process them with additional Mamba blocks, and map the resulting query features to voxel outputs. This branch captures long-range and cross-ROI dependencies.

\subsubsection{ROI Residual Decoder}

The ROI Residual Decoder predicts ROI-wise voxel details through residual learning. For each ROI token, a small MLP estimates voxel residuals relative to the Stage 1 coarse value:
\begin{equation}
  \mathbf{v}_k = r_k + \Delta_k,
\end{equation}
where $r_k$ is the coarse ROI prediction and $\Delta_k$ is the local residual. This formulation uses coarse predictions as priors and focuses local modeling capacity on fine corrections.

\subsubsection{Prediction Fusion and Amplitude Recovery}

The fused voxel prediction combines holistic and residual outputs:
\begin{equation}
  \mathbf{v}_{fused} = \alpha \cdot \mathbf{v}_{global} + (1 - \alpha) \cdot \mathbf{v}_{local},
\end{equation}
where $\alpha$ is a single global learnable parameter, in contrast to Stage 1's input-dependent adaptive gate. This simpler fusion is sufficient because both decoders read from the same shared representation and differ only in decoding granularity (holistic vs. ROI-residual).

To recover appropriate response amplitudes, we apply a learnable scale-recovery layer conditioned on visual features:
\begin{equation}
  s = \mathrm{Sigmoid}(\mathrm{MLP}(\mathbf{h}_{vis})) \cdot \sigma,
\end{equation}
where $\mathbf{h}_{vis}$ is the visual token from the encoded sequence and $\sigma$ is a learnable scale bound. The final prediction is $\hat{\mathbf{v}} = s \cdot \mathbf{v}_{fused}$, allowing the model to adjust output magnitude based on stimulus complexity.

\subsubsection{Training Objective}

The Stage 2 objective balances voxel reconstruction and latent regularization:
\begin{equation}
  \mathcal{L}_{stage2} = \mathcal{L}_{MSE} + \beta \cdot \mathcal{L}_{KL},
\end{equation}
where $\mathcal{L}_{MSE}$ is voxel-level reconstruction loss and
$\mathcal{L}_{KL} = D_{KL}(q(\mathbf{z}|\mathbf{x}) \| p(\mathbf{z}))$ regularizes the posterior toward a standard Gaussian prior. The coefficient $\beta$ controls this trade-off.

\section{Experiments}
\label{sec:experiments}

\subsection{Experimental Setup}

\noindent\textbf{Dataset and Features.} We evaluate our method on the Natural Scenes Dataset (NSD)~\cite{allen2022massive}. Following standard protocols in MindSimulator~\cite{bao2025mindsimulator} and MindEye2~\cite{scotti2024mindeye2}, we use data from subjects 1, 2, 5, and 7. For each subject, we allocate 9,000 image-fMRI pairs for training and 1,000 for testing, focusing on visual cortex regions which yield approximately 15,000 voxels per subject. Visual representations are extracted using a pre-trained DINOv2-ViT-B/14~\cite{oquab2023dinov2} from its final layer, producing a 768-D global CLS token and 256 local patch tokens.

\noindent\textbf{Implementation Details.} CHASMBrain is implemented in PyTorch and trained on a single NVIDIA RTX 4090 GPU. Stage 1 is trained for 150 epochs (batch size 32, AdamW, learning rate $3\times10^{-5}$, Pearson loss weight $\gamma=0.3$, InfoNCE weight $\lambda=0.1$). Stage 2 is trained for 150 epochs (batch size 16, learning rate $10^{-4}$, KL weight $\beta=0.001$, Stage 1 weights frozen). For ROI clustering we set $V = 30$ voxels per cluster, giving $K = 524$ clusters, with top-70\% aggregation. The VAE latent space has dimension 256, injected additively into token representations. The InfoNCE temperature is set to $\tau=0.07$ with in-batch negatives (batch size 32). Full architecture specifications are provided in the Supplementary Material.

\noindent\textbf{Metrics and Baselines.} We evaluate predictive performance using Mean Squared Error (MSE, $\downarrow$) and voxel-wise Pearson Correlation ($\uparrow$), both averaged over all subjects. Following the NSD analysis framework~\cite{allen2022massive}, measurement fidelity is quantified by the noise ceiling estimated from single-trial GLM betas (the refined ``b3'' pipeline). The reported average predictable variance in visual cortex is approximately 36\%, corresponding to a theoretical maximum of Pearson $r\approx0.60$. All baselines use the same subject-wise z-score normalization as our method.

\subsection{Comparison with Baselines}
\label{sec:baselines}

\noindent\textbf{Simple and linear baselines.} All models use the same DINOv2 ViT-B/14 final-layer embedding. The linear baselines collapse spatial structure via global average pooling into a single 768-D vector, whereas CHASMBrain preserves and separately processes the CLS token and 256 patch tokens. The comparison therefore isolates the contribution of preserving and routing this structure through our dual-stream design. Table~\ref{tab:baselines_extended} reports all comparisons at the voxel level (Stage 2 final output).

\begin{table}[ht]
  \centering
  \caption{Voxel-level fMRI prediction performance. All scores are means over subjects 1, 2, 5, 7. Simple baselines and DINOv2 linear probes use the same features as our model; the probe comparison isolates the contribution of the dual-stream and coarse-to-fine design. Simple baselines and probes use the same DINOv2 final-layer features as ours, but with global average pooling. SynBrain and MindSimulator follow their original feature-extraction pipelines ($^\dagger$ is reproduce results); only the dataset, subject split, and z-score normalization are matched.}
  \label{tab:baselines_extended}
  \begin{tabular}{llcc}
    \toprule
    \textbf{Category} & \textbf{Model} & \textbf{MSE} $\downarrow$ & \textbf{Pearson} $\uparrow$ \\
    \midrule
    \multirow{4}{*}{Simple baselines}
      & Ridge Regression (best $\alpha$) & 0.371 & 0.368 \\
      & MLP                             & 0.533 & 0.231 \\
      & GRU                             & 0.461 & 0.237 \\
      & LSTM                            & 0.468 & 0.232 \\
    \midrule
    \multirow{4}{*}{DINOv2 linear probes}
      & Layer 3  & 0.370 & 0.366 \\
      & Layer 6  & 0.359 & 0.403 \\
      & Layer 9  & 0.357 & 0.410 \\
      & Layer 12 & 0.357 & 0.407 \\
    \midrule
    \multirow{2}{*}{Prior works}
      & SynBrain$^\dagger$~\cite{mai2025synbrain} (NeurIPS 2025) & 0.453 & 0.338 \\
      & MindSimulator~\cite{bao2025mindsimulator} (ICLR 2025)  & 0.429 & 0.322 \\
    \midrule
    & \textbf{CHASMBrain (Ours)} & \textbf{0.261} & \textbf{0.429} \\
    \bottomrule
  \end{tabular}
\end{table}

Three findings emerge from Table~\ref{tab:baselines_extended}. \textit{(i)} Plain sequence models (MLP, GRU, LSTM) plateau below ridge regression, confirming that naive temporal modeling is insufficient for fMRI. \textit{(ii)} Linear probes improve monotonically with DINOv2 depth, confirming that semantic depth helps; however, they saturate at layer 12 (Pearson 0.407, MSE 0.357). \textit{(iii)} At the same final-layer feature source, our model achieves a substantial MSE reduction from 0.357 to 0.261 (26.9\%) improvement over the strongest probe
demonstrating that the dual-stream coarse-to-fine architecture contributes meaningfully beyond feature quality alone. Note that Pearson measures only monotonic trend, while the large MSE gap shows our model predicts accurate voxel \textit{amplitudes}, not just correct relative orderings.

\subsection{Qualitative Results}
\label{sec:qualitative}

\begin{figure}[ht]
  \centering
  \includegraphics[width=\textwidth]{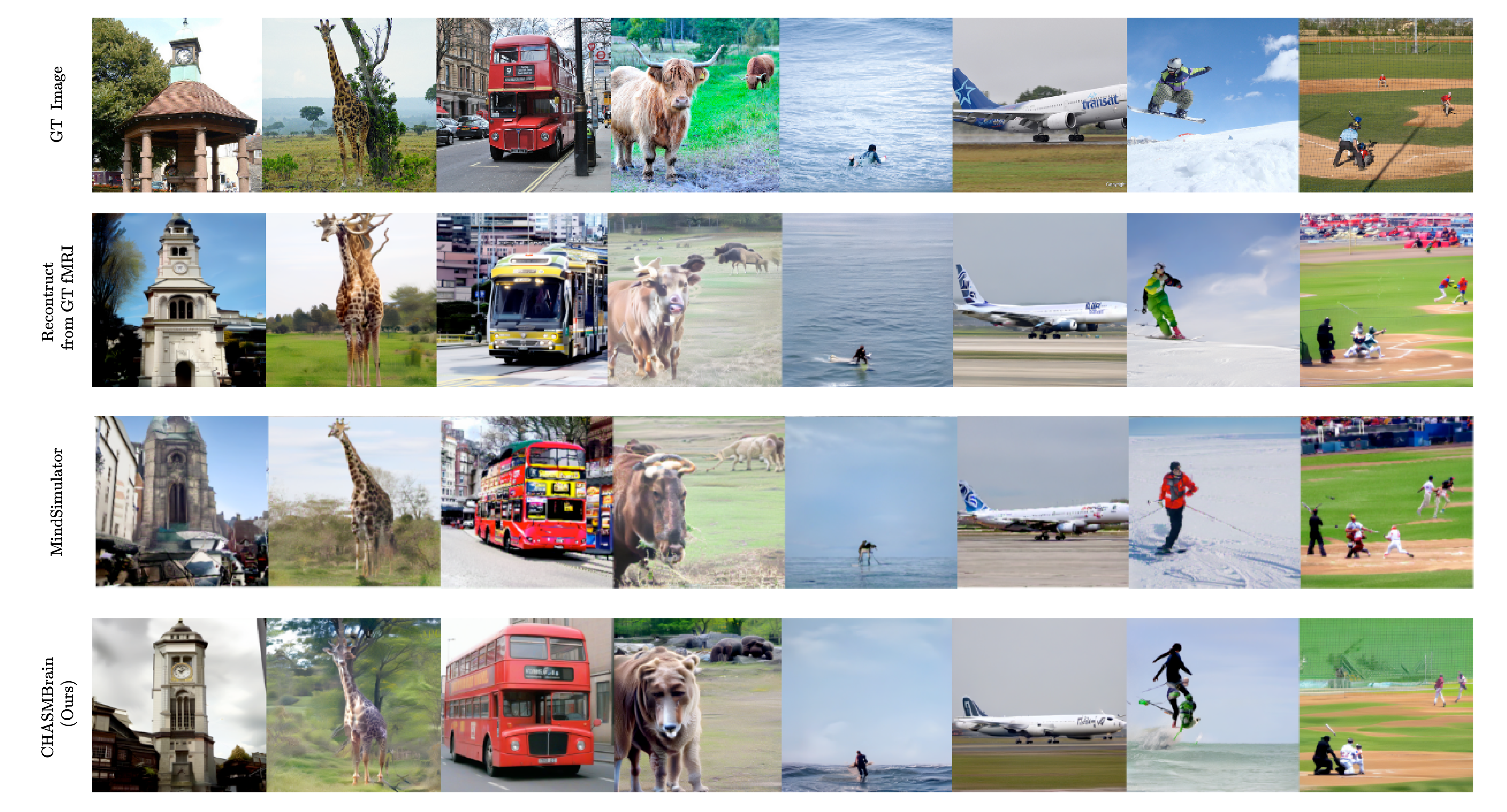}
  \caption{Qualitative comparison of predicted fMRI activations and downstream image reconstructions. CHASMBrain accurately preserves spatial topographic patterns and enables the downstream decoder (MindEyeV2) to reconstruct visual semantics that closely align with the original stimuli.}
  \label{fig:qualitative_comparison}
\end{figure}

\noindent\textbf{Predicted fMRI Maps and Visual Reconstruction.}
To intuitively evaluate our predictions, we analyze both the spatial fMRI activation patterns and their downstream applicability for visual reconstruction (Figure~\ref{fig:qualitative_comparison}).

Mapping the predicted signals back to the cortical surface reveals that CHASMBrain produces accurate spatial activation topologies. Compared to baselines that often yield overly smoothed or scattered activation, our hierarchical approach preserves sharp regional boundaries and localized peaks that closely mirror the ground-truth neural responses. Quantitative reconstruction metrics are provided in the Supplementary Material (Table~S4), where CHASMBrain achieves CLIP similarity 0.849 and DINOv2 similarity 0.558, surpassing MindSimulator on semantic metrics. Notably, it also exceeds reconstructions from ground-truth fMRI on these metrics; we attribute this to our pipeline filtering trial-to-trial noise while preserving semantic structure, rather than to genuinely super-GT signal recovery. This likely reflects a metric–decoder-prior interaction and should be interpreted with caution.

We also feed predicted fMRI signals into an fMRI-to-image decoder (MindEye2) to reconstruct visual stimuli. It is important to acknowledge that visual fidelity is heavily bottlenecked by the decoder's generative priors: even ground-truth fMRI signals fail to perfectly restore original object attributes, establishing a strict visual upper bound. Within these constraints, CHASMBrain demonstrates robust semantic retention---for instance, successfully reconstructing coherent structures of rigid objects (e.g., the red double-decker bus, airplane) significantly better than MindSimulator. Occasional semantic drifts highlight that resolving fine-grained contextual details from predicted fMRI remains challenging.

\begin{figure}[ht]
  \centering
  \includegraphics[width=\textwidth]{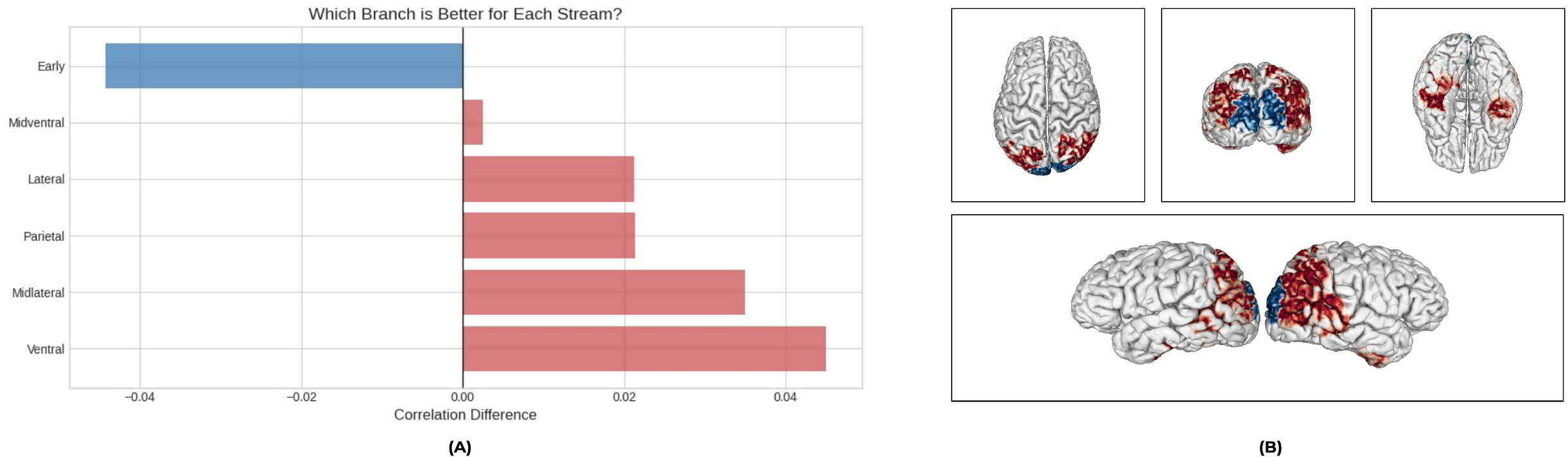}
  \caption{Spatial analysis of dual-stream correlations. \textbf{Left}: Early visual areas (V1/V2) are primarily driven by the Local branch (blue), while higher-order semantic areas are driven by the Global branch (red). \textbf{Right}: 3D visualization showing distinct functional alignments.}
  \label{fig:dual_stream_analysis}
\end{figure}

\noindent\textbf{Dual-Stream Biological Alignment.}
To validate our biological motivation, we map the predictive correlations of the Local (patch) and Global (CLS) branches onto the visual cortex hierarchy (Figure~\ref{fig:dual_stream_analysis}). The Local branch exhibits strong correlations with early visual areas (V1, V2), which are retinotopically organized to process low-level spatial frequencies and edges. The Global branch aligns more strongly with higher-order semantic regions across the lateral and ventral streams. Causal evidence for this specialization is reported in Section~\ref{sec:causal_ablation}.

\subsection{Ablation Studies}
\label{sec:ablation}

\noindent\textbf{Architecture and Refinement Strategy.} Table~\ref{tab:ablation_stages} dissects our architectural components. For Stage 1 coarse prediction (Table~\ref{tab:ablation_stage1}), the proposed Dual-Mamba (Pearson: 0.585) significantly outperforms both a standard Transformer (0.452) and a single-stream Mamba (0.481). Importantly, Dual-Stream Mamba uses \textit{26\% fewer parameters} than the Transformer (14.6M vs.\ 19.8M) and $2\times$ lower FLOPs (2.31G vs.\ 4.52G; see Supplementary Table~S1), ruling out capacity as the source of improvement. The gain stems from the selective state-space design and dual-stream routing.

For Stage 2 voxel refinement (Table~\ref{tab:ablation_stage2}), our VAE-based stochastic refinement achieves the best performance (MSE: 0.261). Diffusion-based refinement fails catastrophically (MSE: 1.973): the forward noise-injection process conflates artificial Gaussian noise with intrinsic fMRI biological noise, severely degrading the signal. In contrast, our VAE smoothly models trial-to-trial variability via KL regularization without explicit noise injection.

\begin{table}[!t]
  \centering
  \caption{Ablation studies: Stage 1 coarse architecture (a) and Stage 2 voxel-level refinement strategy (b). Stage 1 metrics are ROI-level; Stage 2 metrics are voxel-level.}
  \label{tab:ablation_stages}
  \begin{minipage}{0.46\linewidth}
    \centering
    \subcaption{Stage 1: Architecture Design}
    \label{tab:ablation_stage1}
    \begin{tabular}{lcc}
      \toprule
      \textbf{Architecture} & \textbf{MSE} $\downarrow$ & \textbf{Pearson} $\uparrow$ \\
      \midrule
      Transformer         & 0.213 & 0.452 \\
      Mamba (Single)      & 0.156 & 0.481 \\
      \textbf{Dual-Mamba} & \textbf{0.141} & \textbf{0.585} \\
      \bottomrule
    \end{tabular}
  \end{minipage}
  \hspace{0.06\linewidth}
  \begin{minipage}{0.46\linewidth}
    \centering
    \subcaption{Stage 2: Refinement Strategy}
    \label{tab:ablation_stage2}
    \begin{tabular}{lcc}
      \toprule
      \textbf{Method} & \textbf{MSE} $\downarrow$ & \textbf{Pearson} $\uparrow$ \\
      \midrule
      MLP (Baseline)         & 0.535 & 0.391 \\
      w-Dual only            & 0.399 & 0.406 \\
      w-Dual + Diff          & 1.973 & 0.259 \\
      \textbf{w-Dual + VAE}  & \textbf{0.261} & \textbf{0.429} \\
      \bottomrule
    \end{tabular}
  \end{minipage}
\end{table}

\noindent\textbf{Clustering Granularity.} Table~\ref{tab:ablation_clustering} demonstrates the trade-off inherent in ROI clustering. Coarse clustering (100 voxels/cluster) maximizes Stage 1 correlation (0.605) through aggressive denoising but bottlenecks Stage 2 refinement (MSE: 0.561) due to the highly ill-posed one-to-many voxel mapping. Fine clustering (10 voxels/cluster) preserves spatial resolution for Stage 2 but suffers insufficient noise averaging in Stage 1. The empirical optimum at 30 voxels/cluster strikes the best overall balance. The anatomically predefined Schaefer parcellation~\cite{schaefer2018local} performs substantially worse, confirming that flexible data-driven clustering is necessary to capture task-specific visual cortex activations.

\begin{table}[ht]
  \centering
  \caption{Impact of clustering granularity on both prediction stages.}
  \label{tab:ablation_clustering}
  \begin{tabular}{lcccc}
    \toprule
    & \multicolumn{2}{c}{\textbf{Stage 1 (Coarse)}} & \multicolumn{2}{c}{\textbf{Stage 2 (Fine)}} \\
    \cmidrule(lr){2-3} \cmidrule(lr){4-5}
    \textbf{Clustering Strategy} & \textbf{MSE} $\downarrow$ & \textbf{Pearson} $\uparrow$ & \textbf{MSE} $\downarrow$ & \textbf{Pearson} $\uparrow$ \\
    \midrule
    10 voxels/cluster & 0.242 & 0.543 & 0.332 & 0.375 \\
    \textbf{30 voxels/cluster} & \textbf{0.141} & \textbf{0.585} & \textbf{0.261} & \textbf{0.429} \\
    100 voxels/cluster& 0.123 & 0.605 & 0.561 & 0.370 \\
    \midrule
    Schaefer Atlas~\cite{schaefer2018local} & 0.486 & 0.338 & 0.577 & 0.236 \\
    \bottomrule
  \end{tabular}
\end{table}

\noindent\textbf{Top-$p$ threshold sensitivity.} The choice of $p=70\%$ is principled rather than heuristic. At $p=90\%$, Stage 1 ROI Pearson is essentially unchanged (0.586 vs.\ 0.585), but Stage 2 voxel-level MSE degrades by 46\% (0.381 vs.\ 0.261). Noise invisible in the symmetric Stage 1 aggregation task is amplified by Stage 2's one-to-many voxel decoding. The 70\% threshold is the largest value before this amplification appears, making it a principled operating point on the precision--noise trade-off curve.

\subsection{Causal Evidence for Biological Alignment}
\label{sec:causal_ablation}

The spatial correlation analysis in Figure~\ref{fig:dual_stream_analysis} is consistent with the known cortical hierarchy, but correlation alone does not establish that the streams play causally distinct roles. We use ``causal'' throughout in the sense of a controlled intervention on the model's internal wiring at inference time, not as a claim about neural causation in the brain itself. We therefore conduct a causal branch-ablation study in which four variants of the same architecture (identical parameter counts) differ \textit{only} in input routing or branch gating at inference time. Pearson is reported at Stage 1 ROI level, per cortical group (524 ROIs split by NSD visual stream labels: Early, Mid-ventral, Higher).

\begin{table}[ht]
  \centering

  \caption{Voxel-level fMRI prediction performance. All scores are means over subjects 1, 2, 5, 7. Simple baselines and DINOv2 linear probes use the same final-layer features as ours but with global average pooling; this comparison isolates the contribution of the dual-stream and coarse-to-fine design. SynBrain and MindSimulator follow their original feature-extraction pipelines (intrinsic to those models); only the dataset, subject split, and z-score normalization are matched. $^\dagger$ denotes reproduced results.}

  \label{tab:causal_ablation}
  \begin{tabular}{lcccc}
    \toprule
    \textbf{Variant} & \textbf{Early} $\uparrow$ & \textbf{Mid-ventral} $\uparrow$ & \textbf{Higher} $\uparrow$ & \textbf{Overall} $\uparrow$ \\
    \midrule
    Both (full model) & \textbf{0.512} & \textbf{0.589} & \textbf{0.625} & \textbf{0.585} \\
    Where only (patch stream only) & 0.425 & 0.461 & 0.501 & 0.467 \\
    Swap (routes exchanged) & 0.279 & 0.326 & 0.472 & 0.381 \\
    What only (CLS stream only) & 0.273 & 0.313 & 0.446 & 0.364 \\
    \bottomrule
  \end{tabular}
\end{table}

Three findings from Table~\ref{tab:causal_ablation} establish causal rather than correlational alignment. \textit{(i)} Both branches are necessary: removing either stream causes a clear performance drop across all cortical groups. \textit{(ii)} The patch stream specifically carries early-cortex information: removing it (What only) hurts Early regions far more than Higher regions, consistent with retinotopic processing in V1/V2. \textit{(iii)} Routing matters causally: under Swap, with \textit{identical weights}, Early Pearson collapses from 0.512 to 0.279 (45\% relative drop) while Higher is far less affected, from 0.625 to 0.472 (24\% relative drop)---a 1.9$\times$ asymmetry in relative degradation.

These results give a more precise picture than the spatial correlation maps alone: the patch stream is specifically locked to early visual cortex (causally validated), while the CLS stream contributes broader semantic context that benefits higher-order regions. We accordingly refine Contribution~3 (Section~\ref{sec:intro}) to emphasize this asymmetric specialization.

\subsection{Cross-Subject Generalization}
\label{sec:cross_subject}

A legitimate concern for subject-specific fMRI models is whether improvements reflect genuine representational learning or subject-level overfitting. To address this, we train the CHASMBrain backbone on one or more source subjects, freeze it, and fine-tune only a lightweight linear per-subject head (30 epochs, lr $10^{-4}$) on the target subject's training data. Results are reported at voxel level (full pipeline).

\begin{table}[ht]
  \centering
  \caption{Cross-subject transfer. The backbone is trained on source subjects and frozen; only a lightweight per-subject head is fine-tuned on target data. Within-subject (no transfer) is shown for reference.}
  \label{tab:cross_subject}
  \begin{tabular}{lcc}
    \toprule
    \textbf{Source $\rightarrow$ Target} & \textbf{Pearson} $\uparrow$ & \textbf{MSE} $\downarrow$ \\
    \midrule
    subj07 only (within-subject, no transfer) & 0.451 & 0.239 \\
    \midrule
    subj01 $\rightarrow$ subj07           & 0.436 & 0.294 \\
    subj01, 02 $\rightarrow$ subj07       & 0.448 & 0.269 \\
    subj01, 02, 05 $\rightarrow$ subj07   & 0.448 & 0.271 \\
    subj01, 02 $\rightarrow$ subj05       & 0.457 & 0.260 \\
    \bottomrule
  \end{tabular}
\end{table}

The frozen backbone trained on source subjects reaches near within-subject Pearson on the target (e.g., 0.448 vs.\ 0.451 for subj07), despite never observing the target subject during backbone training. Adding more source subjects beyond two yields diminishing returns, suggesting the backbone converges to a shared visual representation. The small remaining gap is attributed to per-subject voxel layout differences, which the lightweight head handles. These results indicate that CHASMBrain's performance improvements are not a product of subject-specific overfitting.

\subsection{Discussion and Limitations}

While CHASMBrain establishes strong performance on NSD, several limitations remain. First, its predictive upper bound is constrained by the inherent physiological noise of fMRI signals (noise ceiling $r\approx0.60$). Second, the pipeline relies on a frozen DINOv2 encoder biased toward static, object-centric images, and may lack temporal dynamics intrinsic to natural human vision. Third, cross-dataset evaluation (e.g., BOLD5000, Algonauts) is left as future work due to differing acquisition protocols and ROI definitions. Finally, downstream qualitative evaluations are bottlenecked by the hallucination tendencies of the fMRI-to-image decoder.

\section{Conclusion}

In this paper, we introduced CHASMBrain, a coarse-to-fine hierarchical architecture for image-to-fMRI encoding. By incorporating a sequential Mamba-based dual-stream design, the model bridges the gap between deep visual features and the hierarchical processing mechanism of the human visual system. The proposed two-stage approach---denoised ROI predictions followed by fine-grained voxel-level refinement via a Mamba-VAE---successfully addresses the challenge of high intrinsic fMRI noise while maintaining spatial and semantic specificity.

Experiments on NSD yield a Pearson correlation of 0.429 and MSE of 0.261, outperforming prior deep encoding models as well as strong linear baselines and DINOv2 layer-wise probes. Beyond predictive metrics, causal branch-ablation experiments reveal an asymmetric specialization: the patch stream is specifically locked to early visual cortex (retinotopic regions), while the CLS stream serves as a broader semantic context for higher-order areas. Cross-subject transfer experiments further show that the learned backbone is subject-agnostic at the representational level, with a lightweight per-subject head sufficient to adapt to individual voxel layouts.

Future work will pursue three directions. First, fine-grained alignment analysis between model components and sub-regional cortical specializations (e.g., face patches, scene-selective regions). Second, extension to multi-modal stimuli---audio, video, and language---to model multi-sensory integration in the brain. Third, cross-dataset evaluation to assess protocol robustness and develop shared-subject encoders that require minimal individual data.

%
%

\bibliographystyle{splncs04}
\bibliography{_main}


\end{document}